\documentclass[11pt,a4paper]{article}
\usepackage{times,latexsym}
\usepackage{url}
\usepackage[T1]{fontenc}
\usepackage[font=small]{caption}

\newcommand{\calX}{\mathcal{X}}

\newcommand{\defeq}{\overset{\mathrm{def}}{=}}

\usepackage[acceptedWithA]{tacl2021v1}
\usepackage{hyperref}

\usepackage{enumitem}
\setlist[itemize]{leftmargin=*}

\usepackage{xspace,mfirstuc,tabulary}

\newif\iftaclinstructions
\taclinstructionsfalse %
\iftaclinstructions
\renewcommand{\confidential}{}
\renewcommand{\anonsubtext}{(No author info supplied here, for consistency with
TACL-submission anonymization requirements)}
\newcommand{\instr}
\fi

\iftaclpubformat %

\else

\fi

\usepackage{amsmath,amssymb,amsfonts,mathrsfs}
\usepackage{wrapfig}
\usepackage{multirow}
\usepackage{xspace}
\usepackage{graphicx}
\usepackage{caption}
\usepackage{subcaption}
\usepackage{booktabs}
\usepackage{bbm}
\usepackage{adjustbox}
\usepackage{scalerel}
\usepackage{makecell}
\usepackage{float}

\newcommand{\bleu}{\textsc{bleu}\xspace}
\newcommand{\xx}{\mathbf{x}}
\newcommand{\yy}{\mathbf{y}}

\newcommand{\calY}{\mathcal{Y}}
\newcommand{\calC}{\mathcal{C}}
\newcommand{\calB}{\mathcal{B}}

\newcommand{\vtheta}{{\boldsymbol \theta}}

\newcommand{\vocab}{\mathcal{V}}
\newcommand{\vocabeos}{\bar{\mathcal{V}}}
\newcommand{\eos}{\textsc{eos}\xspace}
\newcommand{\bos}{\textsc{bos}\xspace}

\usepackage{cleveref}
\crefname{section}{\S}{\S\S}
\Crefname{section}{\S}{\S\S}
\crefformat{section}{\S#2#1#3}
\crefname{figure}{Fig.}{Fig.}
\crefname{alg}{Alg.}{Alg.}
\crefname{line}{line}{lines}
\crefname{appendix}{App.}{}
\crefname{equation}{Eq.}{Eq.}
\crefname{table}{Table}{Tables}
\crefname{prop}{Proposition}{Propositions}

\DeclareMathOperator*{\argmax}{argmax}
\DeclareMathOperator*{\argmin}{argmin}
\usepackage[disable]{todonotes}
\makeatletter
\newcommand*\iftodonotes{\if@todonotes@disabled\expandafter\@secondoftwo\else\expandafter\@firstoftwo\fi}  %
\makeatother
\newcommand{\noindentaftertodo}{\iftodonotes{\noindent}{}}
\newcommand{\note}[4][]{\todo[author=#2,color=#3,size=\scriptsize,fancyline,caption={},#1]{#4}} %

\newcommand{\ryan}[2][]{\note[#1]{ryan}{violet!40}{#2}}
\newcommand{\Ryan}[2][]{\ryan[inline,#1]{#2}\noindentaftertodo}

\newcommand{\clara}[2][]{\note[#1]{clara}{orange!40}{#2}}

\usepackage{inconsolata}

\everypar{\looseness=-1}

\title{On Decoding Strategies for Neural Text Generators}

\author{
Gian Wiher~\;~ Clara Meister~\;~Ryan Cotterell \\
   ETH Z{\"u}rich \\
  \texttt{\href{mailto:gian.wiher@inf.ethz.ch}{gian.wiher@inf.ethz.ch}}~\;~\texttt{\href{mailto:clara.meister@inf.ethz.ch}{clara.meister@inf.ethz.ch}} \\
  \texttt{\href{mailto:ryan.cotterell@inf.ethz.ch}{ryan.cotterell@inf.ethz.ch}}
}

\date{}

\begin{document}
\maketitle
\begin{abstract}{
  When generating text from probabilistic models, the chosen decoding strategy has a profound effect on the resulting text. Yet the properties elicited by various decoding strategies do not always transfer across natural language generation tasks. For example, while mode-seeking methods like beam search perform remarkably well for machine translation, they have been observed to lead to incoherent and repetitive text in story generation. Despite such observations, the effectiveness of decoding strategies is often assessed with respect to only a single task. This work---in contrast---provides a comprehensive analysis of the interaction between language generation tasks and decoding strategies. Specifically, we measure changes in attributes of generated text as a function of both decoding strategy and task using human and automatic evaluation. Our results reveal both previously-observed and surprising findings.  For example, the nature of the diversity-quality trade-off in language generation is very task-specific; the length bias often attributed to beam search is not constant across tasks.\footnote{Code will be released after de-anonymization.}}

\end{abstract}

\section{Introduction}
\begin{figure}[h]
    \centering
    \includegraphics[width=\linewidth]{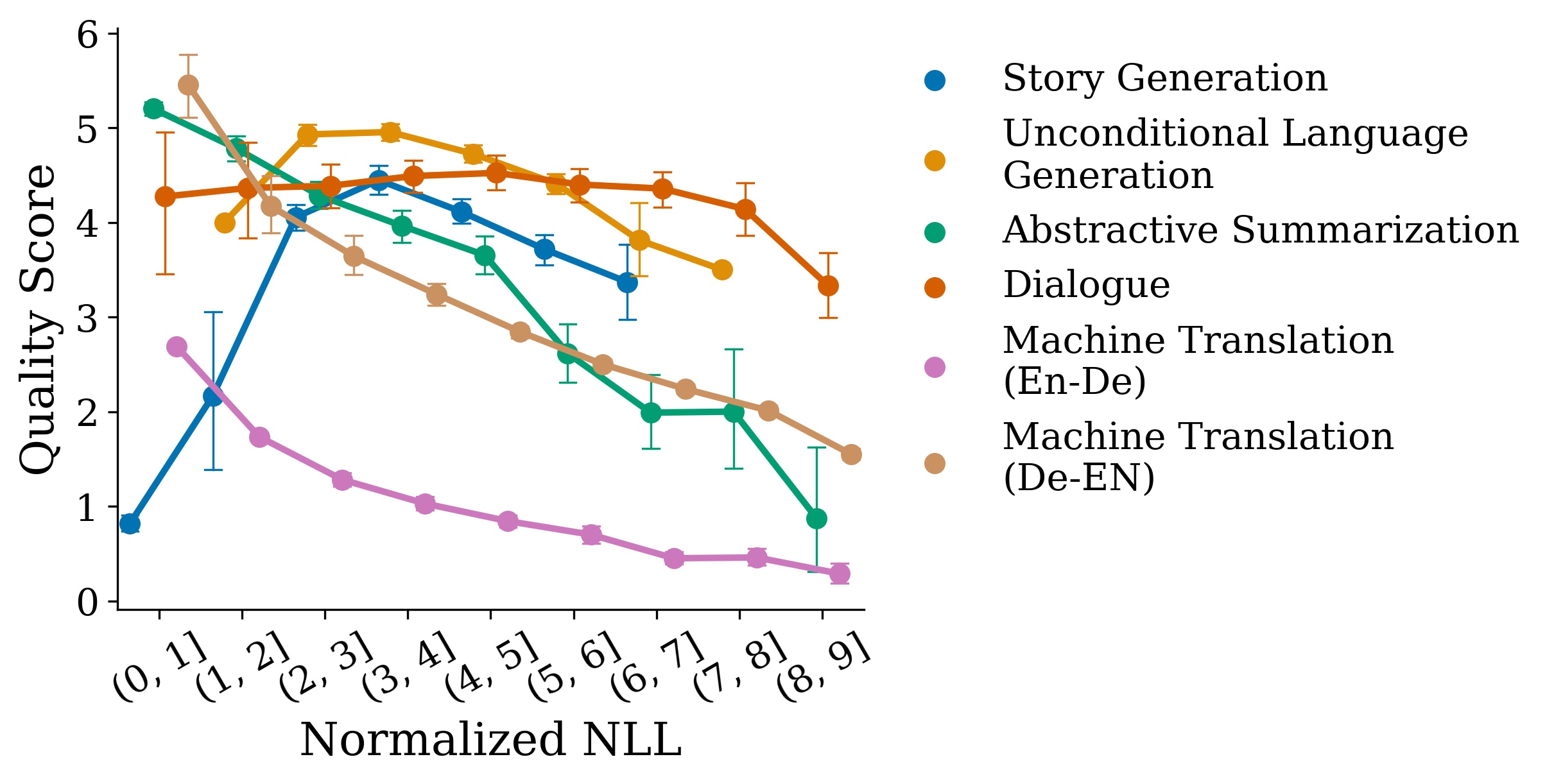}
    \caption{Quality-probability trade-off for different language generation tasks: story generation, unconditional language generation, abstractive summarization, dialogue, and machine translation. Notably, general trends in each curve differ drastically across tasks, despite training models with the same objective. See \cref{sec:diversity} for details on how quality scores are computed.\looseness=-1 
    }\label{fig:quality_probability_tradeoff}
    \vspace{-1em}
\end{figure}

Modern neural networks constitute an exciting new approach for the generation of natural language text.
Much of the initial research into neural text generators went into designing different architectures  \cite{sutskever_seq2seq,rush-etal-2015-neural,serban_dialogue}.
However, recent work has hinted that which \textbf{decoding strategy}, i.e. the method used to generate strings from the model, may be more important than the model architecture itself.
For instance, a well replicated recent result is that, under a probabilistic neural text generator trained with the maximum-likelihood objective, the most probable string is often \emph{not} human-like or high quality \cite{stahlberg_nmt_2019, eikema_is_2020}. %
In light of this finding, a plethora of decoding strategies have been introduced into the literature, each claiming to generate more desirable text than competing approaches.\looseness=-1

Lamentably, empirical studies of decoding strategies are typically evaluated with respect to a \emph{single} natural language generation task---without investigation into how performance may change \emph{across} tasks---despite the fact that these tasks differ qualitatively across a large number of axes. These qualitative differences manifest quantitatively as well: for example, we can see in \cref{fig:quality_probability_tradeoff} that high probability strings are favorable in some tasks, like machine translation (MT), while heavily disfavored in others, like story generation (SG). 
Consequently, we should not \emph{a priori} expect a strategy that works well for one task to demonstrate the same performance in another. 
Indeed, several cases already show evidence of this: Beam search works remarkably well for machine translation 
but outside of this context, has been observed to return dull text or degenerate text \cite{holtzman_curious_2020, delucia_decoding_2020}. This raises a natural fear that decoding strategies have been optimized for performance on a specific task, and the task-agnostic claims about the effectiveness of one decoding strategy over another are potentially ill-founded. A broader analysis of decoding strategies---both within and across tasks---is needed in order to fully understand the extent of such a problem\clara{work on last sentence}.\looseness=-1 

Our work fills this lacuna, providing the first comprehensive comparison of decoding strategies across natural language generation tasks.
Empirically, we compare strategy performance on several axes, taxonomizing methods into groups such as deterministic and stochastic, to understand the importance of various strategy attributes for quantifiable properties of text. \ryan{Rewrite with the same bullet points as letter}
In summary, our main findings include the following:
\begin{itemize}
\setlength{\itemsep}{0em}
    \item Many previous empirical observations, among them the quality-diversity and quality-probability trade-offs \cite{ippolito_comparison_2019,zhang_trading_2020, nadeem_systematic_2020}, manifest themselves in very task-specific ways. For example, our experiments reveal a distinct quality-diversity trade-off albeit only in a certain subset of tasks. 
    This brings into question whether there is a single phenomenon under consideration or many distinct, but related phenomena.\looseness=-1 
    \item A group-level analysis shows the first empirical evidence of a distinct divide in preference for stochastic versus deterministic strategies across tasks: All directed generation tasks appear to favor the latter,\ryan{Break this down into non-technical language. I am not sure what you mean by mode-seeking here.} yet there is a notable trend in the strength of this preference---even the inverse is true for story generation. 
\end{itemize}
We see these results as both a reference point for language generation practitioners, so that they can more confidently choose a decoding strategy that fits their needs, and as an indicator of potential strengths and weaknesses of today's neural probabilistic language generators. We have reason to believe that there is a task-specific optimization happening in the literature whereby many of the proposed and (even celebrated) decoding strategies only outperform their competitors on specific tasks. Thus, our paper, serves as a cautionary note about proper comparisons.

\section{Probabilistic Language Generators}
In this work, we consider models for language generation tasks that define a probability distribution over strings. 
More formally, these models are probability distributions $p$ over an output space $\calY$---(perhaps) conditioned on an input $\xx$---where $\calY$ is the set consisting of all possible strings that can be constructed from the vocabulary $\vocab$:
\begin{equation}
    \calY \defeq \{ \bos \circ \mathbf{v} \circ \eos \mid \mathbf{v} \in \vocab^* \}
\end{equation}
Here, $\bos$ and $\eos$ stand for special reserved beginning-of-sentence and end-of-sentence tokens, respectively, and $\vocab^*$ denotes the Kleene closure of $\vocab$. 

Today's language generators are typically parameterized by encoder-decoder architectures with attention mechanisms \cite{sutskever_seq2seq}, notably the transformer \cite{vaswani_attention_2017}, with trainable weights $\vtheta$. These models follow a local-normalization scheme, meaning that for all $t>0$, $p(\,\cdot \mid \yy_{<t})$ defines a probability distribution over $\vocabeos \defeq \vocab \cup \{\eos\}$. The probability of a sequence $\yy = \langle y_0, y_1, \dots \rangle$ can thus be computed as:\looseness=-1
\begin{equation}
    p(\yy ) = \prod_{t=1}^{|\yy|}p(y_t \mid \yy_{<t})
\end{equation}
where $\yy_{<t} \!\defeq\!\langle y_0, \ldots , y_{t-1}\rangle$ and $\yy_{<1}\!=\!y_0 \!\defeq \!\bos$.\looseness=-1

In order to learn the weights $\vtheta$, we minimize some loss function $L(\vtheta;\calC)$, defined in terms of a corpus $\calC$. In theory, we want examples in $\calC$ to be assigned high probability. Accordingly, our loss is typically their negative log-likelihood under $p$.\footnote{For certain tasks, this loss is typically augmented with the \textbf{label smoothing} \cite{szegedy2016} objective to combat overfitting. In short,  a certain probability mass is discounted from the ground-truth token and redistributed uniformly across all the other tokens.}

\section{The Decoding Problem} \label{decoders}
We define the decoding problem as the search for some string $\yy^*$ according to a given model $p$ and a set of decision rules. 
Given the probabilistic nature of most language generators, the natural choice for such a string would be the most probable sequence under the model\ryan{We might consider adding a log due to convention here}\clara{seeing as we arent doing much with the scoring function, I think that would just be distracting}:
\begin{equation}\label{eq:decodingproblem}
    \yy^\star = \argmax_{\yy \in \calY}\,\, p(\yy \mid \xx)
\end{equation}
Solving the above optimization problem is commonly referred to as maximum a posteriori (MAP) decoding. 
There are two main reasons why in practice this direct optimization is not used when decoding: First, because of the exponentially large space $\calY$ and the non-Markovian structure of commonly used neural generators, direct optimization is often computationally infeasible. Second, recent research has shown that the mode, i.e. the MAP solution $\yy^*$, is often not human-like or high quality text \cite{eikema_is_2020}. For example, in the domain of MT, the most likely string under the model is often the empty string \cite{stahlberg_nmt_2019}. For open-ended generation,\footnote{We define \emph{directed} generation tasks as involving a strong relationship between input and output, e.g., as in MT; for \emph{open-ended} tasks, input contexts only pose a soft constraint on the output space, i.e., there is a considerable degree of freedom in what is a plausible output, e.g. in dialogue or story generation.}  it has been observed that there's a positive correlation between likelihood and quality up to only a certain inflection point, after which the correlation becomes negative \cite{zhang_trading_2020}. Thus in practice $\yy^*$ is almost exclusively approximated using heuristic methods. An overview of such (commonly-used) methods is presented below.\looseness=-1

\subsection{Deterministic Algorithms}
\paragraph{Greedy Search.}
One approximation of $\yy^*$ is obtained by greedily choosing the most probable token at each decoding step $t$, i.e., the following recursion is performed until the \eos symbol is chosen or some maximum time step $T$ is reached:
\begin{align}
    y_0 &= \bos \nonumber\\
    y_t &= \argmax_{y \in \vocabeos}\,\, \log p(y \mid \xx, \yy_{<t}) && \text{
\color{gray}(for $t > 0$)}\nonumber
\end{align}
Note that there is no formal guarantee that greedy decoding will return the global optimum of the decoding objective since decisions are only locally optimal.\looseness=-1

\paragraph{Beam Search.}\label{beamsearch}
Beam search is a simple extension of greedy search.  Rather than considering only the highest probability continuation of our string at each step, we keep the $k \in \mathbb{Z}_+$ highest probability paths, where the hyperparameter $k$ is referred to as the beam:\looseness=-1
\begin{align}
    Y_0 &= \{\bos\}\nonumber \\
    Y_t &= \argmax_{\substack{Y_t' \subseteq \calB_t, \\ |Y_t'| = k}}\ \mathcal{L}(Y'_t) && \text{\color{gray}(for $t > 0$)} \nonumber
\end{align}
where $\calB_t$ is our beam, consisting of all possible extensions of $\yy \in Y_{t-1}$ and $\mathcal{L} \colon \calY \to \mathbb{R}$ is a scoring function  that operates over sets $Y \subseteq \mathcal{Y}$. Typically, we choose $\mathcal{L}(Y) = \sum_{\yy \in Y} \log p(\yy \mid \xx)$. As with greedy decoding, the recursion is performed until all strings end in the \eos symbol or some maximum time step $T$ is reached. The highest scoring string $\yy^*$ is then chosen from this final set $Y_T$.

Other scoring functions have been proposed as modifications to the vanilla beam search algorithm. %
For example, \cite{vijayakumar_diverse_2018} propose \textbf{diverse beam search} (DBS) to address the issue of the lack of diversity within the set of returned strings. The algorithm further splits the beam into several sub-groups and adds an inner iteration at each time step to maximize for inter-group diversity, i.e., they set
\begin{equation}
    \mathcal{L}(Y_t^{(g)}) = \!\!\sum_{\yy \in Y_t^{(g)}}\!\! \log p(\yy \mid \xx) - \lambda\! \sum_{g'< g} \!\Delta(\yy, Y_t^{(g')})
\end{equation}
where $\Delta(\yy, Y_t^{(g)})$ is a measure of dissimilarity between $\yy$ and strings within $Y_t^{(g)}$.\ryan{You need to specify to the similarity function $\Delta$} We refer the reader to the original work for the full algorithm.

\subsection{Stochastic Algorithms}

\paragraph{Ancestral Sampling.}
Instead of approximating $\yy^*$, one can obtain generations by sampling $\yy \sim  p(\cdot \!\mid \xx)$. Due to the local normalization scheme of the models that we consider, this can be achieved simply by setting $y_0 = \bos$ and then drawing each $y_t \sim p(\cdot \!\mid \xx, \yy_{<t})$ until $\eos$ is sampled.\looseness=-1
\begin{table*}
    \centering
    \begin{adjustbox}{max width=\linewidth}
    \begin{tabular}{p{5.5cm}llll}
        \toprule
         Task & $\calX$ & $\calY_{out}$ & $p$ & Dataset  \\
         \hline
         Machine Translation (MT) & sequence in source language & sequence in target language & FAIR’s \textsc{wmt19} submission & \href{http://www.statmt.org/wmt19/metrics-task.html}{\textsc{newstest2019}} \\
         Abstractive Summarization (AS) & news article & summary & BART & \href{https://huggingface.co/datasets/cnn_dailymail}{\textsc{CNN/Dailymail}} \\
         Dialogue (Diag) & conversation history & response & DIALOGPT & \href{https://github.com/microsoft/DialoGPT}{\textsc{DialoGPT}} \\
         Story Generation (SG) & short prompt & related story & GPT-2 (small and medium) & \href{https://github.com/pytorch/fairseq/tree/master/examples/stories}{\textsc{WritingPrompts}} \\
         Unconditional  Generation (ULG) & empty sequence ($\langle\bos\rangle$) & plausible natural language strings & GPT-2 (small and medium)& \href{https://blog.einstein.ai/the-wikitext-long-term-dependency-language-modeling-dataset/#download}{\textsc{WikiText-103}} \\
         \bottomrule
    \end{tabular}
    \end{adjustbox}
    \caption{Overview of the tasks considered in this work. Examples given for $\calX$ and $\calY_{out}$ are the intended input and output, respectively. Models $p$ are evaluated on the test set of the specified dataset. We finetune the GPT-2 models on the specified dataset; other models are loaded from checkpoints provided by the \href{https://huggingface.co/}{Hugging Face} framework \cite{wolf-etal-2020-transformers}.\looseness=-1}
    \vspace{-0.5em}
    \label{tab:tasks}
\end{table*}\todo{if we have space, add citations for each of the datasets later on}
\paragraph{Top-$k$ Sampling.}
Perhaps due to the ``unreliable tail'' of the distribution \cite{holtzman_curious_2020}---i.e., the subset of $\vocabeos$ that are unrealistic extensions of a string but are necessarily assigned probability mass due to the non-sparse nature of the softmax transformation---sampling  directly  from  $ p(\cdot \!\mid\! \xx)$  can lead to text that is  incoherent and sometimes even unrelated to the subject \cite{fan_hierarchical_2018}\clara{ add other citations from before Holtzmann}. One way to  overcome this issue is to limit the sampling space to the top-$k$ most likely tokens in each decoding step. Prior to sampling, the distribution over $\vocabeos$ is recomputed: Let $Z(\xx, \yy_{<t}) \defeq \sum_{y \in {\vocabeos^{(k)}}}^{} p(y \mid \xx, \yy_{<t})$ where $\vocabeos^{(k)} \subseteq \vocabeos$ is defined to be the set of the $k$ most likely tokens. The truncated distribution is given by:
\begin{align}
\pi(y \mid \xx, \yy_{<t}) = 
\begin{cases}
\frac{p(y \mid \xx, \yy_{<t})}{Z(\xx, \yy_{<t})} & \textbf{if } y \in \vocabeos^{(k)} \\
0 & \textbf{otherwise }
\end{cases}\label{eq:topk}
\end{align}

\paragraph{Nucleus (Top-$p$) Sampling.}
Rather than always considering a fixed size set, nucleus sampling dynamically adjusts the number of tokens considered based on the spread of the probability distribution at each generation step.\clara{need motivation? not sure}
Formally, nucleus sampling \cite{holtzman_curious_2020} considers the smallest subset of tokens whose cumulative probability mass exceeds a chosen threshold $p$. 
For generation step $t$, $p \in (0,1]$ and probability distribution $p(\cdot \mid \xx, \yy_{<t})$ let the top-$p$ vocabulary $\vocabeos^{(p)} \subseteq \vocabeos$ be the smallest set such that 
 \begin{align}
     \sum_{y \in \vocabeos^{(p)}}^{}p(y \mid \xx, \yy_{<t}) \geq p
 \end{align}
The truncated distribution is then computed similarly to \cref{eq:topk} with $p' = \sum_{y \in \vocabeos^{(p)}}^{} p(y \mid \xx, \yy_{<t})$.

\paragraph{Bayes Minimum Risk (MBR).} \label{bayes_def}
Under probabilistic language generators, probability mass is often spread over a large set of likely candidates without clear preference \cite{ott2018analyzing}. However, this set of likely strings should not be arbitrary when $p$ is good. Rather, these strings should capture the statistics of training data well, containing a number of potentially good solutions \cite{eikema_is_2020}. This motivates a decision rule that exploits all available information in this set.  Let $u \colon \calY \times \calY \to \mathbb{R}$ be a utility function that evaluates a string $\yy$ against reference $\hat\yy$. According to statistical decision theory \cite{bickel_mathematical_1977}, the optimal decision $\yy^*$ is the one that minimizes expected risk (here we define risk as negative utility):
\begin{align} \label{eq:min_risk}
    \yy^* = \argmin_{\yy \in \calY}\,\,  \mathbb{E}_{p(\hat\yy \mid \xx)}[-u(\yy, \hat\yy)]
\end{align}\\
Like MAP, it is generally computationally infeasible to solve the MBR objective exactly given the size of $\calY$. In practice, one can obtain an unbiased estimate of the expected risk by Monte Carlo (MC) methods and limit the search space for the maximization problem to the sampled set.

\section{Experimental Setup} \label{sec:experiments}
The strategies presented in section \ref{decoders} are compared across a variety of NLG tasks covering open-ended as well as directed generations tasks. We define a task more formally as a triple $(\calX, \calY_{\mathrm{out}}, p)$\ryan{There should be a macro calYout } where $\calX$ denotes the input space, $\calY_{out} \subseteq \calY$ the output space \footnote{Note that formally the input and output spaces only differ by the model specific vocabularies and maximum generation length $l\in \mathbb{Z}_+$.} and $p$ a model that defines a probability distribution over $\calY_{out}$ for every input $\xx \in \calX$. A high-level overview of these tasks (and the respective datasets used) can be found in \cref{tab:tasks}. We use solely transformer-based models, all state-of-the-art for their respective tasks \cite{ng_facebook_2019,lewis_bart_2019,zhang_dialogpt_2020,radford_language_nodate}. We use open-sourced versions of models for reproducibility.
\begin{table*}[h!]
    \centering
    \begin{adjustbox}{max width=\linewidth}
    \begin{tabular}{l>{\footnotesize}p{0.8\linewidth}}
        \toprule
         
         \multicolumn{2}{c}{\textbf{Quality Metrics}} \\ \midrule
          \multicolumn{2}{c}{\small Automatic} \\ \midrule
         
         \textsc{bleu} & Corpus-level metric originally developed to assess translation quality of MT systems \cite{papineni_bleu_2002}. Produces a score between $0$ and $1$ based on modified $n$-gram precision. We use the \textsc{sacrebleu} \cite{post_call_2018} framework.   \\ \hline
         
         \textsc{meteor} & Metric based on the harmonic mean of unigram precision and recall. Originally developed to evaluate MT. We use version $1.5$ of the implementation from \cite{denkowski_meteor_2014}. \\ \hline
         
         \textsc{comet} & Neural framework to train multilingual MT evaluation systems proposed by \cite{rei_comet_2020}. The nature of this metrics makes it only compatible with the MT task.  We use a pretrained model checkpoint  provided by the original work.%
         \\ \hline
         
         \textsc{rouge} & Recall-oriented set of metrics originally developed to assess the quality of automatically generated summaries \cite{lin_rouge_2004}. We report the \textsc{rouge-l} measure, which is based on longest common subsequences between candidate and reference. \\ \hline
         
         \textsc{bleurt} & Trained evaluation metric based on \textsc{bert} \cite{devlin_bert_2019}. Returns a score that indicates to what extent the candidate is grammatical and conveys the meaning of the reference \cite{sellam_bleurt_2020}. We use a pretrained model checkpoint provided by the original work. %
         \\ \midrule
          \multicolumn{2}{c}{\small Human} \\ \midrule
         
         \textsc{adequacy} &  How well does the response/continuation fit in a given conversation history? \\
         \hline
         
         \textsc{naturalness} & To what degree does the text seem to be a natural english text? \\ 
         \hline
         
         \textsc{quality} &  How high is the overall quality of the text? \\
         \hline
         
         \textsc{accuracy} & Given the context, is the text accurate?  \\ \hline
         
         \textsc{fluency} & How fluent is the given text? \\
         \midrule
         \multicolumn{2}{c}{\textbf{Diversity Metrics}} \\ \midrule
         
          \textsc{dist}-$n$ & Number of distinct $n$-grams divided by the total number of $n$-grams \cite{li_diversity-promoting_2016} \\ \hline
          
         \textsc{ent}-$n$ & Entropy of empirical $n$-gram distribution:  $-\sum_{w\in S}\frac{F(w)}{\sum_{w'\in S}^{}F(w')}\log\frac{F(w)}{\sum_{w'\in S}^{}F(w')}$ where $S$ denotes the set of all observed $n$-grams and $F(w)$ is the count of  $w$ \\ \hline
         
         \makecell[l]{$n$-\textsc{gram div.}}  &  Average over dist-$n$ measures for different values of $n$. We calculate the average over $k \in \{ 1,...,5 \}$. \\ \hline
         
         \textsc{self-bleu} & Average \textsc{bleu} score across strings when using all other strings in set as references \cite{zhu_texygen_2018}.\looseness=-1\\ \hline
         
         \textsc{repetition}  &  If a phrase (minimum length of 2) is repeated at least three times until the end of the generation, it is labeled as a repetition. This definition of a repetition is taken from \cite{holtzman_curious_2020} \\

         \bottomrule
    \end{tabular}
    \end{adjustbox}
    \caption{List of metrics considered in this work. For human evaluation metrics, prompt shown is provided to raters.\looseness=-1}
        \vspace{-0.5em}
    \label{tab:qualitymetrics}
\end{table*}\clara{double check human; add in refs to comet and bleurt}

\Ryan{integrate into previous paragraphs}
\paragraph{Decoding Strategy Settings.}\clara{fill out} In this work, we consider beam search with beam sizes $k=5$ and $k=10$, and DBS with Hamming distance as a dissimilarity function, $\lambda=0.7$ and $G=k=5$. The choice of dissimilarity function and hyperparameters is based on the recommendations from the original work. When we only want to return one string, we select the hypothesis with the highest score according to $\log p$. For top-$k$ sampling, we set $k=30$ and for top-$p$ sampling, we set $p=0.85$ based on experiments in \cite{delucia_decoding_2020} that suggest a parameter range $p \in [0.7, 0.9]$ for top-$p$ sampling.    
For MBR, %
we obtain $30$ to $32$ ancestral samples to approximate the expected risk in \cref{eq:min_risk} using Monte Carlo. To speed up the generation process the samples are generated in batches. Depending on the memory requirements of the different models the batch size differ across tasks and we thus have small differences in the number of samples acquired. The candidate sequences, for which we all calculate the expected risk, consists of the ancestral samples used for the Monte Carlo approximations together with sequences obtained from the other decoders. The metric \textsc{BEER} \cite{stanojevic_fitting_2014} is used as utility function $u$.

\subsection{Metrics}\label{sec:metrics}
\begin{figure*}[t]
    \centering
    \includegraphics[width=\linewidth]{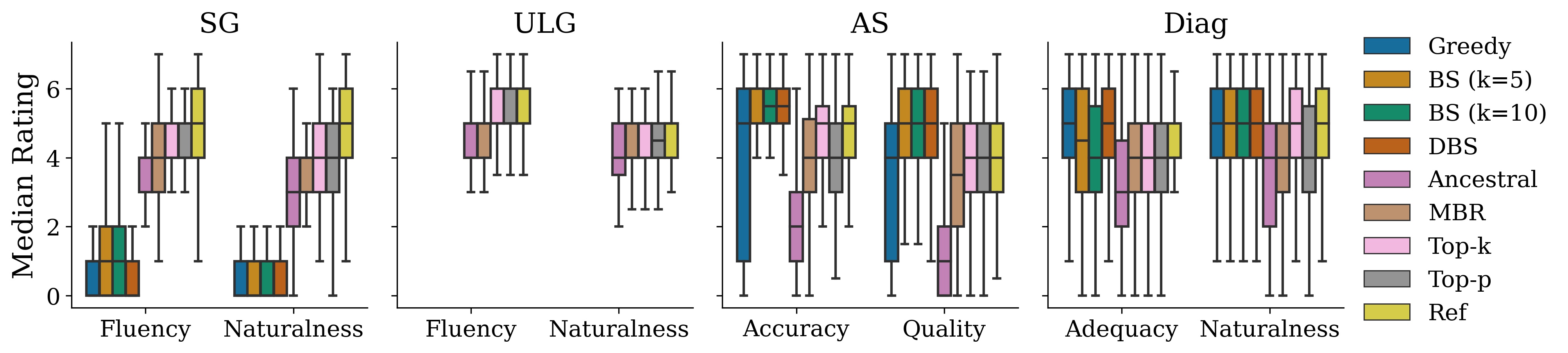}
    \caption{Median human evaluation ratings.}
    \vspace{-0.5em}
    \label{fig:human_eval}
\end{figure*}
We employ a number of different metrics to compare text across decoding strategies.  An overview of all metrics can be found in \cref{tab:qualitymetrics}. Note that we roughly divide the set of metrics into two categories: \emph{diversity} metrics and \emph{quality} metrics. Intuitively, we may expect that the two criteria are not always of equal importance. For example, in MT an accurate, high quality translation of the input is more highly valued than generating engaging or stylized language or a wider range of diverse outputs. On the other hand, 
a conversational agent that is able to talk about a diverse range of topics is likely highly preferred to one that repeats the safest phrases over and over \cite{li_mutual_2016}. In our subsequent experiments, we provide a quantitative analysis of this trade-off.\looseness=-1

\subsubsection{Evaluation of Quality}

For tasks where one has access to a ground truth reference, e.g., MT, AS and to some extent Diag, there are a variety of automatic metrics to evaluate quality. Most of these metrics are based on statistics of $n$-gram overlap between output and reference. This class of metrics has its limitations; consequently we also consider human judgements of text quality using criteria in \cref{tab:qualitymetrics}. We  use the  \href{https://www.prolific.co/}{\emph{prolific}} framework to obtain ratings from $5$ different annotators on $200$ examples per decoding strategy; criteria used for each of the tasks is given in \cref{tab:humancrit}.\footnote{While several works have shown that human evaluations are not always consistent \citep[e.g.,][]{clark-etal-2021-thats}, they are still the standard for evaluating most text generation tasks.} For each of the criteria an $8$-point Likert scale is used. We select the criteria based on which have been most commonly used to assess performance of text generators on a given task, as outlined by \cite{VANDERLEE2021101151}, and describe them to the annotators as in Table \ref{tab:qualitymetrics}. We check the obtained ratings manually if they have been filled out with care. If a rater assigns high scores to multiple examples that do not fulfill the specified criteria at all, the rating is rejected and we obtain a fresh set of scores from a new rater. For SG, AS and Diag the raters are first presented with a prompt/news article/dialogue history followed by the outputs of the different decoders and the reference in random order. For unconditional language generation we present the raters with generations and references in random order. \todo{@Gian, add in details} We omit human annotations for MT because it has been observed that there is no significant gain over the automatic metrics when using crowd workers due to large variations in evaluation \cite{freitag_experts_2021}.

\begin{table}
    \centering
    \begin{adjustbox}{max width=\linewidth}
    \begin{tabular}{l|c|c|c|c|c}
        \toprule
        Task          & \textsc{adequacy}&\textsc{quality}&\textsc{fluency}                          & \textsc{naturalness} &\textsc{accuracy}      \\
        \midrule
        Diag   &   \checkmark &&&\checkmark & \\
        AS     &&   \checkmark &&&\checkmark        \\
        SG      &&&   \checkmark &\checkmark &   \\
        ULG    &&&   \checkmark &\checkmark &  \\
        \bottomrule
    \end{tabular}
    \end{adjustbox}
    \caption{Criteria used for each task in human evals.\looseness=-1}
    \vspace{-1em}
    \label{tab:humancrit}
\end{table}

\subsubsection{Evaluation of Diversity}\label{sec:diversity}
\begin{figure*}[h]
  \centering
  \includegraphics[width=\linewidth]{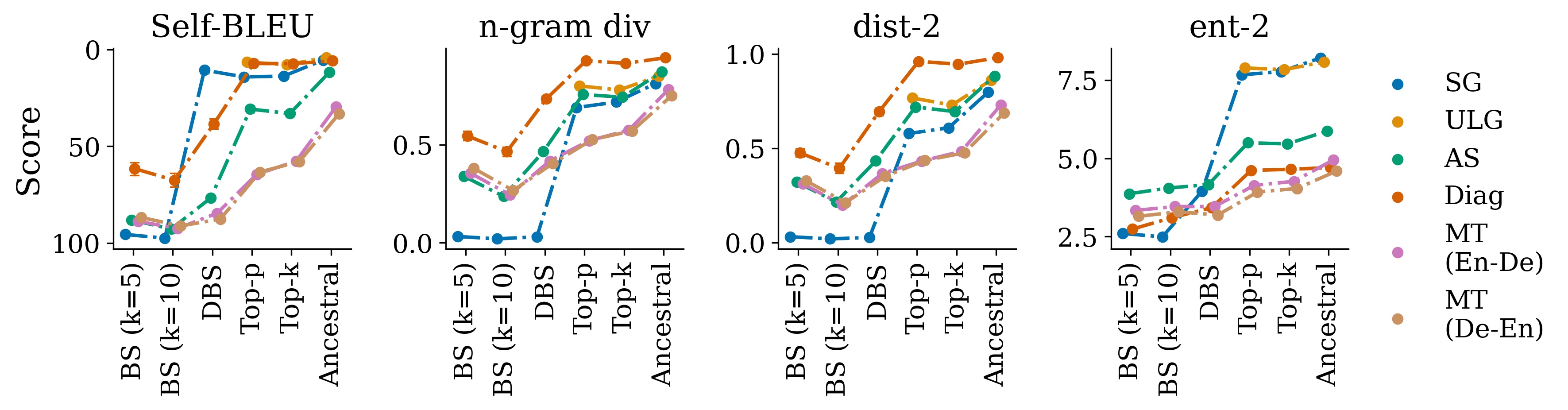} 
\caption{Diversity metrics calculated at the set level. For ULG, the metrics are calculated for randomly chosen (disjoint) subsets of all generations. Note that low self-\bleu indicates high diversity.}
\vspace{-0.5em}
  \label{fig:diversity}
\end{figure*}
Automatic metrics to measure lexical diversity of generated text are mostly based on statistics of $n$-gram counts; while lexical diversity is a narrow definition of diversity, it is the commonly employed one in language generation as diverse word choice is arguably a large factor for this characteristic. Note that lexical diversity can be measured at the string level, i.e., within a given string $\yy$, or at the set level, i.e. across different strings $\{ \yy^{(1)},\yy^{(2)}, 
\dots\}$. While we provide some results for the former set of metrics, we focus largely on the latter set, as often practitioners are more concerned with having a diverse set of generations \emph{per} input. Specifically, we take measurements with respect to sets decoded by each strategy, i.e., the size $K$ set decoded by beam search or $K$ items generated according to a specific stochastic scheme.\footnote{Because greedy and MBR decoding are methods that only return a single string, they are not considered in the latter set of metrics.} For the beam search methods $K$ is equal to the beam size while for the sampling decoders we set $K = 10$. For each input $\xx \in \calX$ we thus obtain a set of strings per decoder over which we calculate various metrics such as self-\bleu or $n$-gram diversity. Self-\bleu is calculated as the average of \bleu scores when setting one of the generations $x_i \in \calX$ as candidate and all other strings in $\calX$ as references. To calculate dist-$n$, ent-$n$, and $n$-gram diversity metrics for a set of generations, we concatenate all strings in $\calX$ and perform calculations as described in \cref{tab:qualitymetrics}. For ULG, where we only have one input $\xx$, we instead calculate scores over random (disjoint) subsets of size $K=10$.\looseness=-1

\section{Results}
\subsection{Quality}
Human evaluations are aggregated across raters, using the median value for each string. Results are displayed in  \cref{fig:human_eval}. 
According to human raters, sampling directly from the model yields text with the lowest quality metrics across all tasks: the clear exception is for SG, where we observe that mode-seeking strategies lead to degenerate text (further discussion in \cref{sec:corrs}). In general, for the directed generation tasks (AS and Diag), beam search variants perform the best, even outperforming human generated references. Interestingly, despite its limited exploration of the search space, greedy decoding generates texts on par with beam search methods for Diag.\looseness=-1 %

On the other hand, the results of stochastic methods are more nuanced: while top-$p$ and top-$k$ decoding generate more highly-rated texts than ancestral sampling, they often fail to reach quality levels of the beam search based methods.  MBR decoding, which perhaps falls somewhere between the classes of deterministic and stochastic, likewise performs somewhere in between these classes in terms of quality metrics. Overwhelmingly, trends in performance are much more distinct when analyzing strategies as stochastic vs. deterministic, rather than individually, suggesting that small algorithmic differences in decoding strategies may not be as critical as prior work has made seem \cite{holtzman_curious_2020}.\clara{other citations?}

We present automatic quality evaluation metrics for directed generation tasks in \cref{tab:corpus_eval_1}---the number in brackets in \cref{tab:corpus_eval_1} shows how many of the decoders performed worse than the best one, as determined by a permutation test. We use a significance level of $0.01$; the resulting $p$-values were corrected for multiple testing using a Bonferroni correction.. We observe similar trends as with our human evaluations: beam search methods perform best, followed by top-$p$ and top-$k$ sampling with ancestral sampling performing worst. Despite mixed results in \cref{fig:human_eval}, MBR decoding yields competitive results in terms of automatic evaluation metrics, even matching the performance of beam search; this is perhaps not surprising given the poor correlation between human and automatic evaluation that is frequently observed in language generation. On Diag we see that for the metrics considered we only observe a significant difference in performance between the best decoder and the worst 3, resp worst 4 decoders. Similarly, on MT, we observe that except for the \bleu metric, only a significant difference between the best and the worst 3 decoders is present. On the other hand we have that for the AS the best performing decoder significantly outperforms any other decoder except the other beam search methods. This contrasts the observation for Diag and MT where the mode seeking decoders seem all to perform equal. \todo{discuss significance of results, i.e., where the results of permutation tests are significant}\looseness=-1

\begin{table*}[h]
    \centering
    \small
    \begin{adjustbox}{max width=\textwidth}
    \footnotesize
    \begin{tabular}{lllllllllllll}
    \toprule
    {} & \multicolumn{3}{c}{Dialogue} & \multicolumn{3}{c}{Summarization} & \multicolumn{3}{c}{MT (De-En)} & \multicolumn{3}{c}{MT (En-De)} \\
    {} & \textsc{human} &    \textsc{bleu} &    \textsc{rouge-l}&    \textsc{human} &      \textsc{bleu} &    \textsc{rouge-l} &    \textsc{bleu} &    \textsc{meteor}& \textsc{comet} &   \textsc{bleu} &    \textsc{meteor} &\textsc{comet}  \\
    \midrule
    Greedy  &  4.660 &   0.661 &  $\textbf{9.072}_{(4)}$ &    3.671 &    16.560 &  28.027 &                       40.083 &  42.444 &  0.548 &                       42.072 &  59.174 &  0.613\\
    BS ($k=5$)  & 4.495 &   $\textbf{0.758}_{(3)}$ &  8.796 &    $\textbf{5.235}_{(5)}$ &   17.197 &  31.138  &                       41.049 &  43.005 &  $\textbf{0.561}_{(3)}$ &                       $\textbf{42.746}_{(5)}$ &  $\textbf{59.602}_{(3)}$ &  0.622 \\
    BS ($k=10$)  & 4.456 &    0.746 &  8.331 &   5.180 &     16.726 &  30.650 &                       $\textbf{41.211}_{(5)}$ &  $\textbf{43.101}_{(3)}$ &  0.560 &                       42.680 &  59.583 &  $\textbf{0.622}_{(3)}$  \\
    DBS & $\textbf{4.689}_{(3)}$ &   0.436 &  8.708 &  5.122 &       $\textbf{18.141}_{(5)}$ &  $\textbf{31.487}_{(6)}$ & 39.770 &  42.254 &  0.538 &                       41.702 &  58.793 &  0.611\\
    MBR & 3.815 &    0.510 &  8.469 &    3.709 &    10.771 &  25.120 &                       40.811 &  42.952 &  0.547 &                       42.370 &  59.241 &  0.605\\
    Ancestral  & 3.329 &    0.196 &  5.408 &    1.825 &    5.390 &  17.985  &                       17.402 &  27.425 & -0.520 &                       15.595 &  35.722 & -0.832\\
    Top-$k$   & 4.234 &   0.308 &  6.961 &   4.276 &     11.644 &  25.961 &                       27.574 & 35.651 &  0.376 &                       27.091 &  47.839 &  0.458\\
    top-$p$   & 3.914 &    0.308 &  6.331 & 3.976 &        11.785 &  25.505 &                       29.397 &  36.704 &  0.382 &                       29.998 &  49.778 &  0.481 \\
    \bottomrule
    \end{tabular}
    \end{adjustbox}
    \caption{Corpus-level quality metrics for Diag, AS, and MT. For Diag and AS the human score is calculated by taking the mean over the two criteria upon which the text is rated.}
    \label{tab:corpus_eval_1}
\end{table*} 

\subsection{Diversity}
\begin{figure}[h]
    \centering
    \includegraphics[width=\linewidth]{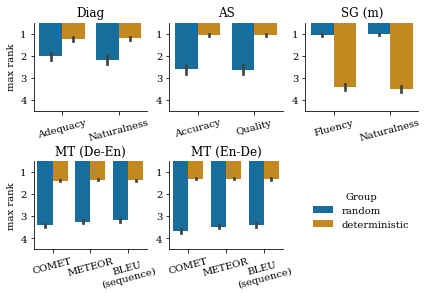}
    \caption{Highest ranks achieved by stochastic vs. deterministic strategies on each input; a rank of 1 means a generation from the respective group of decoding strategies was ranked $1$st among all generations. We omit ULG since only stochastic strategies are considered for this task. Note that the lowest possible rank for a deterministic strategy is $4$ and for a stochastic strategy is $5$.\looseness=-1}
    \vspace{-0.5em}
    \label{fig:randomvsdeterministic}
\end{figure}
We report diversity metrics for different strategies and tasks in \cref{fig:diversity}. Points are connected to better illustrate general trends across diversity metrics, not due to a quantitative relationship between the metrics themselves. We see that in general, the trends for a given task are quite consistent across diversity metrics, i.e., lines of the same color follow the same trend. On the other hand, trends \emph{across} tasks are not as similar. For example, the gap in diversity between deterministic and stochastic methods is much more exacerbated in SG than MT. 

Across tasks, ancestral sampling consistently produces the most diverse outputs. Limiting the search space, as in top-$k$ and top-$p$ sampling, leads to a drop in diversity compared to pure sampling; notably, this drop appears to be much more significant for directed generations tasks. Interestingly, introducing a diversity promoting term, as in DBS, increases diversity with respect to beam-based decoding algorithms, but still leads to substantially less diverse strings than stochastic methods.  

At the task-level, responses for Diag seem to be more inherently diverse than for other tasks. Even methods known for producing repetitive sets, e.g., beam search, generate a relatively diverse set of solutions. This suggests that even though diverse options are often desired in Diag, we may not need to explicitly optimize for them via the chosen decoding strategy.  On the other hand, diversity in SG is quite sensitive to the chosen decoding strategy, displaying drastic differences.\looseness=-1
\todo{@Gian, can you check on the caption for the diversity fig?}

\subsection{Quantitative Trade-offs in NLG Tasks}\label{sec:tradeoffs}\clara{could also name this section 'tradeoffs'}
We provide an analysis of the importance of different metrics for each of the language generation tasks, looking specifically at their relationships with perceived quality.

\paragraph{The Probability--Quality Relationship.} Natural language generation is performed almost solely using probabilistic models. While ideally, we would like high quality text to be assigned high probability (and vice versa), we see that in practice this is not always the case \cite{pmlr-v97-cohen19a,stahlberg_nmt_2019, holtzman_curious_2020,zhang_trading_2020,delucia_decoding_2020}. 
The trends observed in \cref{fig:quality_probability_tradeoff} reveal that while high probability is often a determinant of quality in directed generation tasks, such as MT and AS,\footnote{As computational constraints make it difficult (if not infeasible) to decode the highest probability string from neural models, we do not observe behavior at the extreme end of \cref{fig:quality_probability_tradeoff}, which other works have observed to produce poor quality text.}  there is a negative correlation between quality and probability in SG and ULG at least up until a certain inflection point. 
Such relationships have been a main motivation behind research into new decoding strategies \citep[e.g., ][]{li_diversity-promoting_2016,Shao2017GeneratingLA,holtzman_curious_2020}.

\begin{figure*}[t!]
  \centering
  \includegraphics[width=\linewidth]{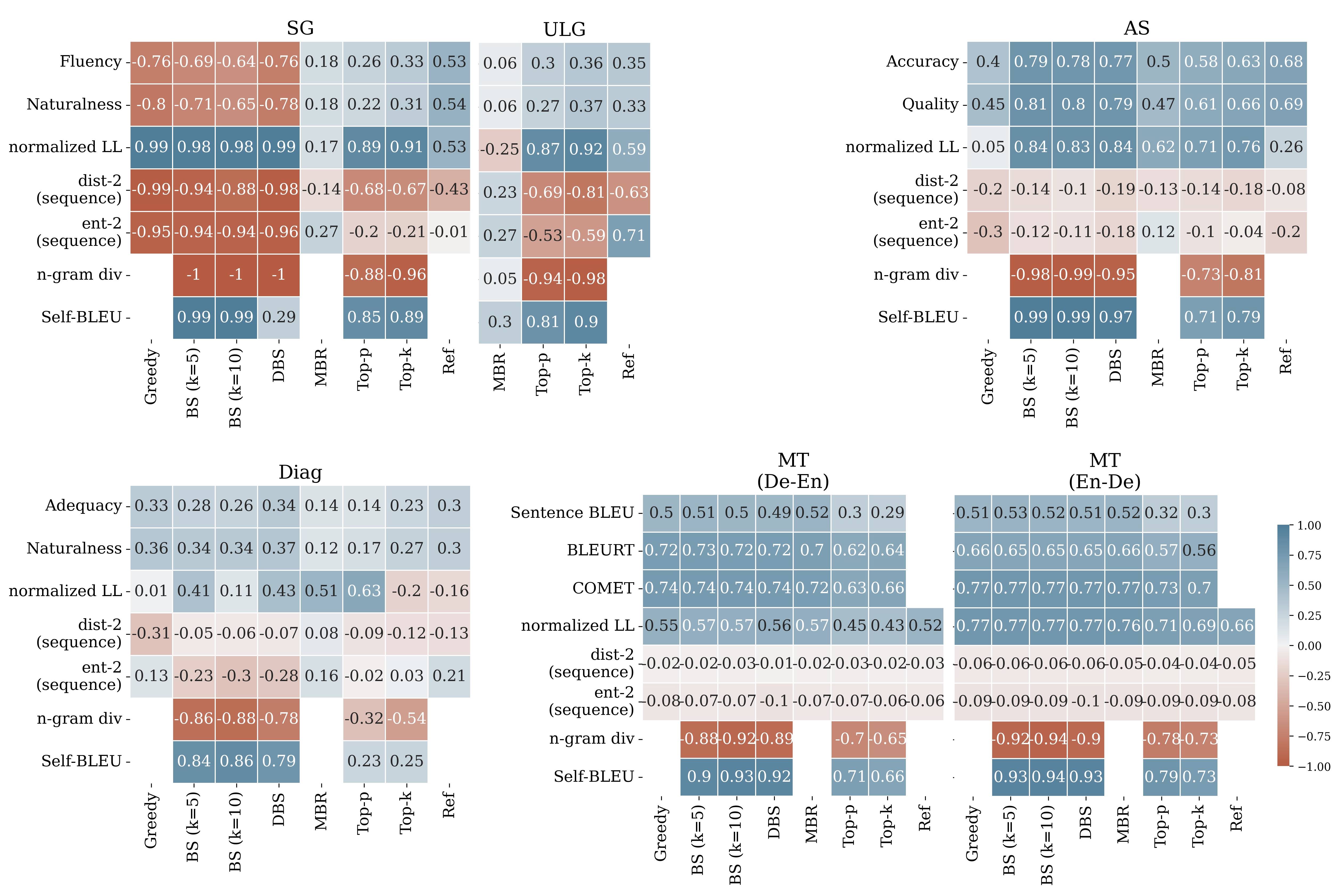} 
\caption{Correlations between quality metrics and other quantitative attributes of text with different decoding schemes separated by task. For each decoder we concatenate the generations with ancestral samples and calculate the Person correlation between the metrics and a one-hot encoding indicating whether or not the generation is an ancestral sample or not.}
\vspace{-0.5em}
  \label{fig:corrs}
\end{figure*}

This relationship also manifests in the divide in performance between deterministic strategies---all of which to some extent are mode-seeking---and stochastic strategies. Naturally, deterministic decoding strategies produce (on average) higher probability strings, as  probability is part of the algorithms' objectives. \cref{fig:corrs} shows that when compared to ancestral samples, most beam search generations are more strongly associated with higher (length) normalized log-likelihood than the output of the sampling based decoders. Thus, we might expect the results observed in \cref{fig:quality_probability_tradeoff} to  appear in a comparison of deterministic and stochastic strategies. 
We rank strategies within a task according to human ratings when available and calculate the highest rank obtained by each of the two groups. More specifically, for each input, we order generations according to their median human rating. Ranks are then assigned to each decoding strategy according to this ordering (lower is better). We then look at the highest rank achieved by the two subsets of decoding strategies.\looseness=-1%

From \cref{fig:randomvsdeterministic} we can see a distinct divide in preference for deterministic vs. stochastic strategies across tasks: all directed generation tasks appear to favor mode-seeking strategies. Yet there is a notable trend in the strength of this preference. As we might intuitively expect, we see an upward trend in the difference in rankings of mode-seeking vs. stochastic decoding methods as a task becomes more semantically-constrained. At one end of the spectrum, in SG, we observe that in nearly all cases, the most highly ranked output from a deterministic strategy is still ranked below the worst of the stochastic strategies,\footnote{This must be the case since the average maximum ranking for mode-seeking methods is almost $4$.\looseness=-1} indicating the ill-suitedness of mode-seeking strategies for such tasks. The opposite is true of MT at the other end of the spectrum.\looseness=-1

\begin{figure}
    \centering
    \includegraphics[width=\linewidth]{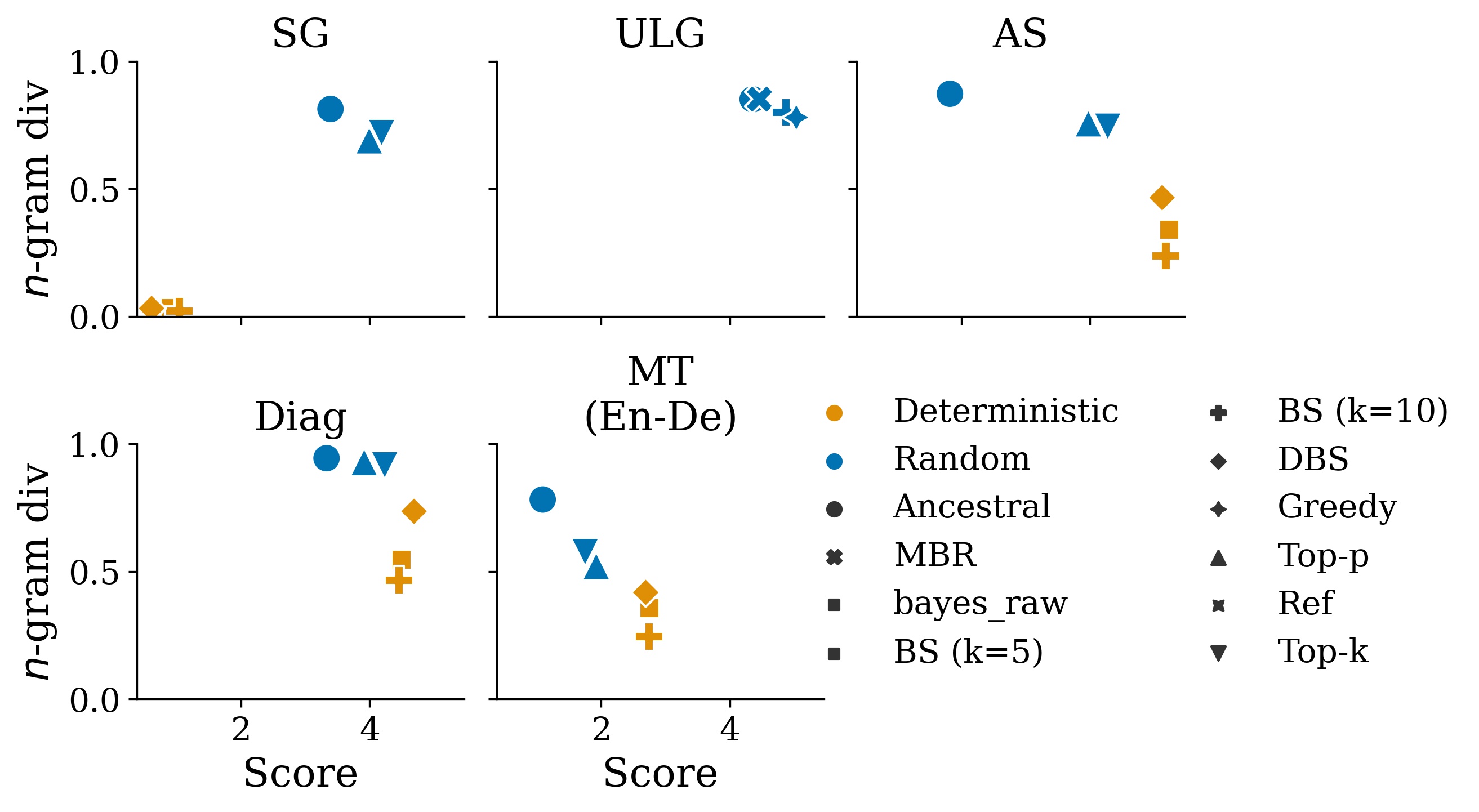}
    \caption{The relationship between diversity ($n$-gram div) and quality (median human rating) across language generation tasks. Results are qualitatively the same when using other diversity metrics, as we might expect given results in \cref{fig:diversity}.}
    \vspace{-0.5em}
    \label{fig:quality_diversity_tradeoff}
\end{figure}

\paragraph{The Diversity--Quality Relationship.} Here we investigate how diversity---as quantified by metrics in \cref{sec:metrics}---relates to quality in a given task. Note that the probability--quality relationship has previously been attributed to a \emph{trade-off} between diversity and quality \cite{zhang_trading_2020,nadeem_systematic_2020}, albeit only in the investigation of a small subset of language generation tasks. However, we see in \cref{fig:quality_diversity_tradeoff} that the relationship between diversity and probability is not so easily defined: it changes quite drastically across tasks.  

Specifically, \cref{fig:quality_diversity_tradeoff} shows there is indeed a trade-off for the two quantities in AS and MT, yet there appears to be an \emph{interdependence} for open-ended generation tasks. Notably, Diag appears to fall outside of this paradigm, which perhaps challenges its definition as a directed generation task. In conjunction with other results, e.g., \cref{fig:corrs}, the trends shown in \cref{fig:quality_diversity_tradeoff} suggest that within directed tasks, Diag falls closer to open-ended generation tasks on the task spectrum. 
We further see that stochastic and deterministic methods are distinctly divided along the diversity--quality trend in each task; although this result is perhaps expected, the separating line is surprisingly sharp in all cases.\looseness=-1

\subsection{Eliciting Quantitative and Qualitative Metrics }\label{sec:corrs}
We now look at the ability of different  decoding strategies to elicit the qualitative metrics described in \cref{sec:metrics}, the quantitative properties studied in \cref{sec:tradeoffs}, as well as certain undesirable attributes of text. Through this analysis, we hope to ascertain how the effectiveness of different decoding strategies generalizes across tasks, and which---if any---more general claims can be made about these strategies.

\cref{fig:corrs} shows how different decoders correlate with various metrics, using ancestral samples as a baseline.\footnote{Ancestral samples give us an unbiased sample of the type of text that is assigned probability mass by our model, thus making it a good baseline for observing the effects of decoding strategies.} Our first take-away is that, these correlation plots differ notably across tasks, which further demonstrates the sensitivity of the performance of decoding strategies to the task at hand. Among these differences though, we observe certain trends that provide insights into how decoders' abilities to generate certain types of texts transfers across tasks. For example, the performance of decoders \emph{within} the subsets of directed and open-ended generation tasks is reasonably consistent. We first discuss more specific trends with respect to \emph{quality} metrics.

\paragraph{Quality Metrics.}
We first note that there is no single decoding method that consistently correlates most strongly with high-quality text, which heads further warnings against more general claims made about decoder performance. 
Perhaps the most distinct result when looking at decoders' correlations with quality metrics is the difference in correlations for mode-seeking methods between open-ended and directed generation tasks. Here we see that on the directed tasks, the use of mode-seeking methods appears to correlate highly with quality, with no substantial differences among this class of methods even when e.g., also optimizing for intra-set diversity (as in DBS). Interestingly, the strengths of the correlations shown by stochastic methods are much more consistent across all tasks than the mode-seeking methods. 

While in general, decoder performance w.r.t. quality metrics is relatively consistent for directed generation tasks, there are exceptions to this consistency: MBR correlates well with quality metrics for MT, but underperforms in comparison to other decoders for both AS and Diag. On AS, greedy search tends to lead to poorer quality text than top-$p$ and top-$k$ sampling where for the other directed tasks, all mode-seeking methods provide generations of higher quality.\looseness=-1

\begin{figure}[t]
    \centering
    \includegraphics[width=\linewidth]{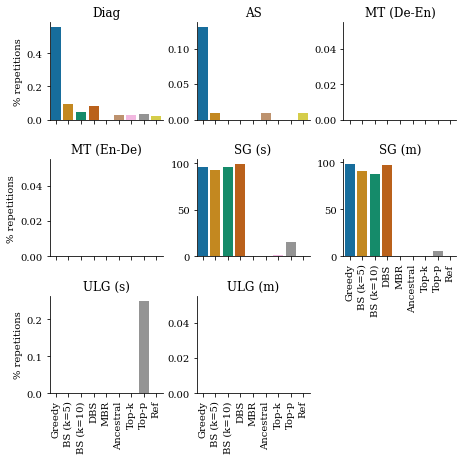}
    \caption{Fraction of generations that degenerate into repetition (see \cref{tab:qualitymetrics} for definition). Note the different scales for the different tasks.}
    \label{fig:repetition}
\end{figure}

\paragraph{Diversity Metrics.}
In comparison to quality, we observe that behavior of different decoders changes less with respect to diversity. 
For diversity metrics calculated over a set of generations, ancestral sampling consistently generates the most diverse text (as demonstrated by the negative $n$-gram diversity/positive self-\bleu correlations shown by all decoders). This is true even in comparison to DBS, which optimizes for intra-set diversity.\footnote{Although in general, DBS seems to be relatively effective at optimizing for intra-set diversity in comparison to other decoders, even achieving low self-\bleu on SG despite also causing degenerations, as shown in \cref{fig:repetition}. } Mode-seeking decoding strategies consistently have a stronger negative correlation with set-level diversity metrics, e.g., self-\bleu, than their stochastic counterparts. This difference is more pronounced on certain tasks: for example, both \cref{fig:corrs} and \cref{fig:diversity} show a bigger jump in diversity scores between DBS and top-$p$ sampling on SG compared to MT or AS. 
Interestingly, there is little consistency across tasks in terms of sequence-level string diversity.

\paragraph{Repetitions.} Probabilistic language generators are known to occasionally produce text with degenerate qualities \cite{dinan, holtzman_curious_2020, welleck2019neural}. One common form of degenerate behavior is repetitions, where generation falls into a loop of repeating the same phrase until the the decoding algorithm terminates. Here we analyze the fraction of times this behavior occurs for different strategies; results can be found in \cref{fig:repetition}.
 On the SG task, we observe a substantial amount of text degeneration for mode-seeking strategies; this holds true for both small and medium variants of GPT-2.  Across both open-ended tasks, the only stochastic decoding scheme that appears to elicit this degenerate behavior is top-$p$ sampling; while only a small percentage of samples, it is responsible for all of the degenerate behavior observed for the ULG task. 
 \begin{table}[h]
    \centering
    
    \begin{adjustbox}{max width=\linewidth}
    \begin{tabular}{lrrrr}
    \toprule
    {} &  \multicolumn{2}{l}{Story Gen. (small)} &  \multicolumn{2}{l}{Story Gen. (medium)} \\
    {}             &  \% repetition  &  ppl                & \% repetition &    ppl        \\
    \midrule
    Greedy              &        95.67&         1.07 &        98.47 &              1.09 \\
    BS ($k=5$)   &             92.58 &         1.11 &      90.70&                 1.11 \\
    BS ($k=10$)  &             95.67&         1.11 &       88.01&                1.11 \\
    DBS &             99.25&         1.05 &        97.75&               1.05 \\
    MBR                 &             0.20&        27.19 &        0&             28.46 \\
    Ancestral  &             0.23&        30.43 &          0.13&            32.98 \\
    Top-$k$               &             1.97&         7.10 &           0.53&            7.38 \\
    Top-$p$             &             15.87&         5.52 &           5.65&            6.33 \\
    Reference           &             0&        23.83 &           0&           19.28 \\
    \bottomrule
    \end{tabular}
    \end{adjustbox}
    \caption{Perplexities and repetition count for different strategies on the SG task. Mode-seeking strategies are able to produce text with very low perplexity but these generations almost always degenerate into repetitions.}
    \vspace{-0.5em}
    \label{tab:ppl}
\end{table}

Notably, for all tasks besides SG, we see repetitive behaviour in less than $1\%$ of generations.%
The exact repetition counts together with the perplexity of the generated texts for SG are shown in \cref{tab:ppl}.

\paragraph{Length.} 
\begin{figure}
    \centering
    \includegraphics[width=\linewidth]{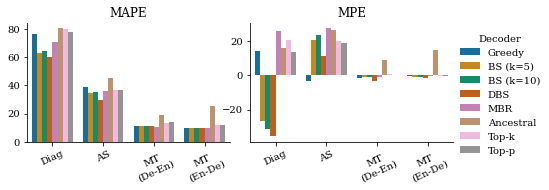}
    \caption{Differences between lengths of generated texts to reference strings. MAPE denotes the mean absolute percentage difference between reference lengths and the lengths of generated texts. MPE denotes the mean percentage error, where we do not take the absolute value of the difference in lengths, in order to get a sense of whether generated strings are (on average) longer or shorter than the reference.}
    \vspace{-0.5em}
    \label{fig:lengths_directed}
\end{figure}

\begin{table}
    \centering
    \small
    \begin{adjustbox}{max width=0.6\linewidth}
    \begin{tabular}{lrr}
    \toprule
    {} & SG  & ULG \\
    \midrule
    Greedy               &                 921.96 & - \\
    BS ($k=5$)   &                       287.11 & -\\
    BS ($k=10$)  &                           247.89 & -\\
    DBS  &                917.73 & -\\
    MBR   &                     681.94 & 450.90\\
    Ancestral            &                    578.57 & 440.56 \\
    Top-$k$                &                        610.73 & 469.99\\
    Top-$p$              &               667.85 &  458.32 \\
    \bottomrule
    \end{tabular}
    
    \end{adjustbox}
    \caption{Mean lengths of generated text for open-ended tasks. Results are displayed from models based on the medium-sized version of GPT-2; we omit results for the small version, which were ostensibly the same.}
    \vspace{-0.5em}
    \label{tab:lenghts_stories}
\end{table}
We further investigate how different decoding strategies affect the length of  generated text. Length biases have frequently been observed in language generation tasks \cite{murray-chiang-2018-correcting,welleck-etal-2020-consistency}, both for shorter and longer strings. In this experiment, we hope to observe how much the decoding scheme can be held responsible for these biases. We report results in \cref{fig:lengths_directed} and \cref{tab:lenghts_stories}. For MT, all strategies manage to generate strings of lengths similar to the reference with the exception of ancestral sampling, which produces slightly longer strings. Interestingly, there are no consistent trends for beam search variants across the other directed generation tasks; rather, trends seem to be inverted for Diag and AS.\looseness=-1 

We see large variation in the length of generated strings for the SG task, especially among mode-seeking strategies; for example, standard beam search produces rather short strings while DBS and greedy decoding produce inordinately long strings. For the unconditional language generation, task we observe no big differences in generated sequence length among stochastic methods. Collectively, these results tell us that the previously observed length biases are task--decoder specific, rather than purely decoder specific.

\section{Discussion}
When constructing a text generation pipeline, the choice of decoding strategy has a large effect on various aspects of the resulting text. Yet when making this choice for a specific language generation task, practitioners are currently limited to either basing their decision on non-comprehensive analyses, using expensive human annotations or even resorting to guesswork. There are potential pitfalls in these practices: as evidenced by various results in this work, certain properties of decoding schemes---especially quality---do not transfer across tasks. This work aims to provide guidance for practitioners in the choice of decoding strategies, revealing their strengths and weaknesses with respect to individual tasks while also giving insights into whether one can expect these properties to transfer to tasks outside of this study. 
While all of the takeaways from this work cannot be summarized in a few lines, we highlight some key observations below.

The relationships and trade-offs between certain properties of text changes notably from one task to another. For example, as depicted in \cref{fig:quality_probability_tradeoff}, high-probability strings are typically also of high quality for MT while there is an almost inverse relationship between these attributes for SG. As shown in \cref{fig:quality_diversity_tradeoff}, a quality--diversity trade-off exists for directed generation tasks whereas for open-ended generation tasks, the relationship is almost a co-dependence. These task-specific characteristics must be taken into account when both choosing and developing decoding strategies.  

While decoder performance generally does not transfer faithfully across  tasks, we can still identify some rules from our experiments that practitioners can use. For one, we see that on directed generation tasks, mode-seeking methods all perform competitively in terms of quality. Further, for stochastic decoders, we observe that restricting the sample space---as done in top-$p$ and top-$k$ decoding---greatly increases quality compared to ancestral sampling, albeit sacrificing some diversity. The ability of a decoder to elicit diversity in text---at least at the set-level---is perhaps the most consistent decoder quality across tasks. There are many other use-case specific insights that can be drawn from the results shown by figures and statistics in this work, which we hope serve as further guidance for practitioners. 

It is worth noting that the behavior of decoders depends on their respective hyperparameters, e.g., $k$ or $p$ in top-$k$ and top-$p$ sampling. This work does not perform a thorough search over hyperparameters, instead employing those most widely-used in order to optimize for the usefulness of our results to practitioners, who are likely to use similar default settings. While based on the results of other works, these choices should provide representative variants of the text generated according to the respective decoding strategy, this is a limitation of our work worth considering.

\section{Conclusion}
This work provides an extensive analysis of the effects of different decoding strategies across various language generation tasks. We show how different attributes of model-generated text change depending not just on decoding strategy, but also on the task at hand,  using both human and automatic evaluations. Our results both confirm several prior observations, e.g., a trade-off between diversity and quality metrics for specific NLG tasks, while also revealing a number of previously unobserved trends in language generation, both with respect to decoding strategies and the tasks themselves. A main take-away of these results is that decoding strategies are perhaps optimized for specific language generation tasks and that practitioners should take great care in basing their choice of decoding strategy off of results reported for alternate tasks. 
We release the evaluation framework and generations in the hopes that this type of analysis will be extended, e.g., by ablating components of model or training strategies, in order to isolate which artefacts can be attributed to the nature of a specific generation task vs. design choices.
We ultimately see this 
line of research as important for both helping practitioners more confidently choose a decoding strategy that fits their needs, without the use of valuable resources, for the further development of decoding strategies and for better understanding the shortcomings of probabilistic language generators.

\bibliography{tacl2018}
\bibliographystyle{acl_natbib}

\end{document}